\documentclass{article}
\usepackage{amsmath,epsfig,bm, amssymb,subfigure,graphicx,algorithm,algorithmic,color}
\newtheorem{defi}{Definition}

\newtheorem{lemm}[defi]{Lemma}

\newcommand{\proofend}{\hfill$\Box$\vspace{2mm}}

\newcommand{\argmin}{\mathop{\mathrm{argmin\,}}}

\newcommand{\mathbbR}{\mathbb{R}}

\newcommand{\boldzero}{{\boldsymbol{0}}}
\newcommand{\boldone}{{\boldsymbol{1}}}

\newcommand{\boldA}{{\boldsymbol{A}}}

\newcommand{\boldC}{{\boldsymbol{C}}}
\newcommand{\boldD}{{\boldsymbol{D}}}

\newcommand{\boldG}{{\boldsymbol{G}}}
\newcommand{\boldH}{{\boldsymbol{H}}}
\newcommand{\boldI}{{\boldsymbol{I}}}

\newcommand{\boldM}{{\boldsymbol{M}}}

\newcommand{\boldP}{{\boldsymbol{P}}}

\newcommand{\boldR}{{\boldsymbol{R}}}

\newcommand{\boldU}{{\boldsymbol{U}}}
\newcommand{\boldV}{{\boldsymbol{V}}}
\newcommand{\boldW}{{\boldsymbol{W}}}
\newcommand{\boldX}{{\boldsymbol{X}}}

\newcommand{\boldZ}{{\boldsymbol{Z}}}

\newcommand{\bolde}{{\boldsymbol{e}}}
\newcommand{\boldf}{{\boldsymbol{f}}}
\newcommand{\boldg}{{\boldsymbol{g}}}

\newcommand{\boldp}{{\boldsymbol{p}}}

\newcommand{\boldu}{{\boldsymbol{u}}}

\newcommand{\boldw}{{\boldsymbol{w}}}
\newcommand{\boldx}{{\boldsymbol{x}}}
\newcommand{\boldy}{{\boldsymbol{y}}}
\newcommand{\boldz}{{\boldsymbol{z}}}

\newcommand{\boldlambda}{{\boldsymbol{\lambda}}}

\newcommand{\calM}{{\mathcal{M}}}
\newcommand{\calN}{{\mathcal{N}}}

\newcommand{\calP}{{\mathcal{P}}}

\newcommand{\calX}{{\mathcal{X}}}
\newcommand{\calY}{{\mathcal{Y}}}

\usepackage{url,multirow}

\advance\oddsidemargin-1.0in
\date{\today}
\title{Convex Factorization Machine for Regression} 

\author{Makoto Yamada$^{1,2}$,  Wenzhao Lian$^3$, Amit Goyal$^2$, Jianhui Chen$^2$, Kishan Wimalawarne$^1$,\\
 Suleiman A Khan$^4$, Samuel Kaski$^{1,4}$, Hiroshi Mamitsuka$^1$, Yi Chang$^2$\\
$^1$Kyoto University, Kyoto, Japan\\
$^2$Yahoo Labs,  CA, USA\\
$^3$Duke University, NC, USA\\
$^4$Aalto University, Espoo, Finland\\
\texttt{makoto.m.yamada@ieee.org}}

\begin{document}
\maketitle

\begin{abstract}
 We propose the \emph{convex factorization machine} (CFM),  which is a convex variant of the widely used Factorization Machines (FMs). Specifically, we employ a linear+quadratic model and regularize the linear term with the $\ell_2$-regularizer and the quadratic term with the \emph{trace norm} regularizer. Then, we formulate the CFM optimization as a semidefinite programming problem and propose an efficient optimization procedure with Hazan's algorithm. A key advantage of CFM over existing FMs is that it can find a globally optimal solution, while FMs may get a poor locally optimal solution since the objective function of FMs is non-convex. In addition, the proposed algorithm is simple yet effective and can be implemented easily. Finally, CFM is a general factorization method and can also be used for other factorization problems including multi-view matrix factorization and tensor completion problems. Through synthetic and movielens datasets, we first show that the proposed CFM achieves results competitive  to FMs. Furthermore, in a toxicogenomics prediction task, we show that CFM outperforms a state-of-the-art tensor factorization method.
\end{abstract}

\section{Introduction}
In recommendation task including movie recommendation and news article recommendation, the data are represented in a matrix form, $\boldA \in \mathbbR^{|U| \times |I|}$, where $\boldA$ is extremely sparse. Matrix factorization (MF), which imputes  missing entries of a matrix with the \emph{low-rank} constraint, is widely used in recommendation systems for  news recommendation, protein-protein interaction prediction, transfer learning, social media user modeling, multi-view learning, and modeling text document collections,  among others \cite{koren2009matrix,wu2007collaborative,xu2003document,pan2010transfer,xu2010protein,hong2013co,virtanen2011bayesian,lian2015integrating,gunasekarconsistent,yan2015scalable}. 

Recently, a general framework of MF called the \emph{factorization machines} (FMs) has been proposed \cite{rendle2010factorization,rendle2012factorization,rendle2013scaling}. FMs are applied to many regression and classification problems, including the display advertising challenge\footnote{\url{https://www.kaggle.com/c/criteo-display-ad-challenge}}, and they show state-of-the-art performance. The key contribution of the FMs is that they reformulate recommendation problems as regression problems, where the input $\boldx$ is a feature vector that indicates the $k$-th user and the $k'$-th item, and output $y$ is the rating of the user-item pair:
\begin{align*}
\boldx_{i} &= [\overbrace{0~\cdots 0~\underbrace{1}_{k\text{-th user}}~0~\cdots 0}^{|U|}~\overbrace{0 ~\cdots 0~\underbrace{1}_{k'\text{-th item}}~0 \cdots 0}^{|I|}]^\top \in \mathbbR^{d}, \\
y_i &= [\boldA]_{k,k'}.
\end{align*}
Here, $d = |U|+ |I|$ is the dimensionality of $\boldx$, $[\boldA]_{k,k'}$ is the score of the $k$-th user and $k'$-th item, and $|\boldA| = n$ is the number of non-zero elements. The goal of the FMs is to find a model that predicts $y$ given an input $\boldx$.

For FMs, the following linear + feature interaction model is employed:
\begin{align*}
f(\boldx;\boldw, \boldG) = w_0 + \boldw_0^\top \boldx + \sum_{\ell = 1}^d \sum_{\ell' = \ell + 1}^d \boldg_\ell^\top \boldg_{\ell'} x_\ell x_{\ell'},
\end{align*}
where $w_0 \in \mathbbR$, $\boldw_0 \in \mathbbR^d$, and $\boldG = [\boldg_1, \ldots, \boldg_m] \in \mathbbR^{d \times m}$ are model parameters ($m \ll d)$. Since only the $k$-th user and $k'$-th item element of the input vector $\boldx$ is non-zero, the model can also be written as
\begin{align*}
\widehat{\boldA}_{k,k'} = w_0 + [\boldw_{0}]_{k}  + [\boldw_{0}]_{|U| + k'} + \boldg_{k}^\top \boldg_{|U|+k'},
\end{align*}
which is equivalent to the matrix factorization model with global, user, and item biases. Moreover, since FMs solve the matrix completion problem through regression, it is easy to utilize side information such as about user's and article's meta information by simply concatenating the meta-information to $\boldx$.

For regression problems, the model parameters are estimated by solving the following optimization problem:
\begin{align*}
\min_{w_0,\boldw, \boldG} &\hspace{0.1cm} \sum_{i = 1}^n (y_i \!-\! f(\boldx_i;w_0,\boldw_0, \boldG))^2 \\
&\hspace{0.1cm}\!+\! \lambda_1 \|w_0\|_2^2 + \lambda_2\|\boldw_0\|_2^2 + \lambda_3\|\boldG\|_{F}^2,
\end{align*}
where the $\lambda_1$, $\lambda_2$, and $\lambda_3$ are regularization parameters, and $\|\boldG\|_F$ is the Frobenius norm.  In \cite{rendle2012factorization}, stochastic gradient descent (SGD), alternating least squares (ALS), and Markov Chain Monte Carlo (MCMC) based approaches were proposed.  These optimization approaches work well in practice if regularization parameters and the initial solution of  parameters are set appropriately. However, since the loss function is non-convex with respect to $\boldG$, it can converge to a poor local optimum (mode). The MCMC-based approach tends to obtain a better solution than ALS and SGD. However, it requires running the sampler long enough to explore different local modes. 

In this paper, we propose the \emph{convex factorization machine} (CFM). We employ the linear+quadratic model, Eq.~\eqref{cfm-model} and estimate $\boldw$ and $\boldW$ such that the squared loss between the output $\boldy$ and the model prediction is minimized. More specifically, we regularize the linear parameter $\boldw$ with the $\ell_2$-regularizer and the quadratic parameter $\boldW$ with the \emph{trace norm} regularizer.  Then, we formulate the CFM optimization problem as a semidefinite programming problem and solve it with Hazan's algorithm  \cite{hazan2008sparse}, which is a Frank-Wolfe algorithm \cite{frank1956algorithm,jaggi2013revisiting}.  A key advantage of the proposed method over existing FMs is that CFM can find a globally optimal solution, while FM can get poor locally optimal solutions. Moreover, our proposed CFM framework is a general variant of convex matrix factorization with nuclear norm regularization, and the CFM algorithm is simple and can be implemented easily. Finally, since CFM is a general factorization framework, it can be easily applied to any factorization problems including multi-view factorization problems \cite{khan2014bayesian}. We demostrate the effectiveness of the proposed method first through synthetic and real-world datasets. Then, we show that the proposed method outperforms a state-of-the-art multi-view factorization method on toxicogenomics data.

\vspace{.1in}
\noindent{\bf Contribution:} The contributions of this paper are summarized below:
\begin{itemize}
\item We formulate the FM problem as a semidefinite programming problem, which is a convex formulation.
\item We show that the proposed CFM framework includes the matrix factorization with nuclear norm regularization \cite{jaggi2010simple} as a special case.
\item We formulate a Tucker based tensor completion problem \cite{tucker1966some,tomioka2010estimation,tomioka2013convex} as a CFM problem. Thanks to the formulation, we can naturally handle large-scale sparse tensor completion problems. To our knowledge, this is the first work. 
\item We propose a simple yet efficient optimization procedure for the semidefinite programming problem using Hazan's algorithm \cite{hazan2008sparse}. 
\item We applied the proposed CFM for a toxicogenomics prediction task; it outperformed a state-of-the-art method.
\end{itemize}

\section{Proposed Method}
In this section, we propose the \emph{convex factorization machine} (CFM) for regression problems.
\subsection{Problem Formulation}
We suppose that we are given $n$ independent and identically distributed (i.i.d.) paired samples $\{(\boldx_i, y_i)~|~\boldx_i \in \calX,~~y_i \in \calY,~i=1,\ldots,n\}$ drawn from a joint distribution with density $p(\boldx, y)$. We denote $\boldX = [\boldx_1, \ldots, \boldx_n] \in \mathbbR^{d \times n}$ as the input data and $\boldy = [y_1, \ldots, y_n]^\top \in \mathbbR^{n}$ as the output real-valued vector. 

The goal of this paper is to find a model that predicts $y$ given an input $\boldx$.  

\subsection{Model}

We employ the following model:
\begin{align}
\label{cfm-model}
f(\boldx;\boldw, \boldW) &= w_0 + \boldw_0^\top \boldx + \sum_{\ell=1}^d \sum_{\ell'= \ell + 1}^d [\boldW]_{\ell,\ell'} x_\ell x_{\ell'},\nonumber \\
&= \boldw^\top \boldz \!+\! \frac{1}{2} \text{tr}( \boldW (\boldx \boldx^\top \!-\! \text{diag}(\boldx \circ \boldx))),
\end{align}
where $\boldz = [1~ \boldx^\top]^\top \in \mathbbR^{d+1}$, $\boldw = [w_0 ~ \boldw_0]^\top \in \mathbbR^{d+1}$, $\boldW \in \mathbbR^{d \times d}$  is a positive semi-definite matrix,  $\text{tr}(\boldX)$ is the trace operator, $\circ$ is the elementwise product, and $\text{diag}(\boldx) \in \mathbbR^{d \times d}$ is the diagonal matrix.  The difference between the FMs model and Eq.~\eqref{cfm-model} is that $\boldg_k^\top \boldg_{k'}$ is parametrized as $W_{k,k'}$. 

The model can equivalently be written as 
\begin{align*}
f(\boldx;\boldw, \boldW) = [\boldw^\top~ \text{vec}(\boldW)^\top] \left[
\begin{array}{c}
\boldz \\
\frac{1}{2}\text{vec}(\boldx \boldx^\top \!-\! \text{diag}(\boldx \circ \boldx))\\
\end{array}
\right],
\end{align*}
where $\text{vec}(\boldW) \in \mathbbR^{d^2}$ is the vectorization operator. Since the model is a linear model, the optimization problem is jointly convex with respect to both $\boldw$ and $\boldW$ if we employ a loss function such as squared loss and logistic loss.

\subsection{Optimization problem}
We formulate the optimization problem of CFM as a semidefinite programming problem:
\begin{align}
\label{eq:CFM-optimization}
\min_{\boldw, \boldW} &\hspace{0.3cm}  \|\boldy -  \boldf(\boldX; \boldw,\boldW)\|_2^2 + \lambda_1 \|\boldw\|_2^2 \nonumber \\
 \text{s.t.} &\hspace{0.3cm}  \boldW \succeq 0~\text{and}~\|\boldW\|_{\text{tr}}  =  \eta, 
\end{align}
where 
\[
\boldf(\boldX; \boldw,\boldW) = [f(\boldx_1;\boldw,\boldW), \ldots, f(\boldx_n;\boldw,\boldW)]^\top \in \mathbbR^{n},
\]
 and $\lambda_1 \geq 0$ and $\eta \geq 0$ are regularizaiton parameters. 
$\|\boldW\|_{\text{tr}}$ is the trace norm  defined as
\[
\|\boldW\|_{\text{tr}} = \text{tr}(\sqrt{\boldW^\top \boldW}) = \sum_{i = 1}^d \sigma_i,
\]
where $\sigma_i$ is the $i$-th singular value of $\boldW$. The trace norm is also referred to as the \emph{nuclear norm} \cite{combettes2005signal}.  Since the singular values are non-negative, the trace norm can be regarded as the $\ell_1$ norm on singular values. Thus, by imposing the trace norm, we can make $\boldW$ to be low-rank. 

To derive a simple yet effective optimization algorithm, we first eliminate $\boldw$ from the optimization problem Eq.\eqref{eq:CFM-optimization} and convert the problem to a convex optimization problem with respect to $\boldW$. Specifically, we take the derivative of the objective function with respect to $\boldw$ and obtain an  analytical solution for $\boldw$:
\begin{align*}
\boldw^{\ast}  &= (\boldZ \boldZ^\top + \lambda_1 \boldI_{d+1})^{-1} \boldZ(\boldy - \boldf_Q(\boldX;\boldW)),
\end{align*}
where 
\begin{align*}
\boldf_Q(\boldX;\boldW) &= [f_Q(\boldx_1;\boldW),\ldots, f_Q(\boldx_n;\boldW)]^\top \in \mathbbR^n,\\
f_Q(\boldx;\boldW) &= \frac{1}{2}\text{tr}( \boldW (\boldx\boldx^\top - \text{diag}(\boldx\circ\boldx))),
\end{align*}
is the model corresponding to the quadratic term of $f(\boldx;\boldw,\boldW)$ such that $f(\boldx;\boldw,\boldW) = \boldw^\top \boldz + f_Q(\boldx;\boldW)$,  $\boldZ = [\boldz_1, \ldots, \boldz_n] \in \mathbbR^{d+1 \times n}$,  $\boldI_{d+1} \in \mathbbR^{d+1 \times d+1}$ is the identity matrix. Note that, $\boldw^\ast$ depends on the unknown parameter $\boldW$. 

Plugging $\boldw^{\ast}$ back into the objective function of Eq.\eqref{eq:CFM-optimization}, we can rewrite the objective function as
\begin{align}
\label{eq:CFM-optimization-2}
\min_{\boldW} &\hspace{0.1cm}  J(\boldW)~~\text{s.t.}~~ \boldW \succeq 0~\text{and}~\|\boldW\|_{\text{tr}} = \eta,
\end{align}
where 
\[
J(\boldW) = (\boldy - \boldf_Q(\boldX;\boldW))^\top \boldC (\boldy - \boldf_Q(\boldX;\boldW)),
\]
$\boldC = \boldR^\top \boldR + \lambda_1 \boldH^\top \boldH$,  $\boldR = \boldI_n - \boldZ^\top (\boldZ \boldZ^\top + \lambda_1 \boldI_{d+1})^{-1}\boldZ$, and $\boldH =  (\boldZ \boldZ^\top + \lambda_1 \boldI_{d+1})^{-1}\boldZ$. 

Once $\widehat{\boldW}$ is obtained by solving Eq.~\eqref{eq:CFM-optimization-2}, we can get the estimated linear parameter $\widehat{\boldw}$ as 
\begin{align*}
\widehat{\boldw}  &= (\boldZ \boldZ^\top + \lambda_1 \boldI_{d+1})^{-1} \boldZ(\boldy - \boldf_Q(\boldX;\widehat{\boldW})).
\end{align*}

\vspace{.1in}
\noindent {\bf Relation to Matrix Factorization with Nuclear Norm Regularization:}
The constraint on $\boldW$ can be written as
\begin{align*}
\boldW = \left[
\begin{array}{cc}
\boldU \boldU^\top & \boldM \\
\boldM^\top & \boldV \boldV^\top \\
\end{array}
\right]\succeq 0,
~\text{tr}(\boldU\boldU^\top) + \text{tr}(\boldV\boldV^\top) = \eta,
\end{align*}
where $\boldU \in \mathbbR^{|U| \times m}$, $\boldV \in \mathbbR^{|I| \times m}$, and $\boldM = \boldU\boldV^\top \in \mathbbR^{|U| \times |I|}$. Furthermore, for the CFM setting, the $k$-th user and $k'$-th item rating is modeled as
\begin{align*}
[\widehat{\boldA}]_{k,k'} = w_0 + [\boldw_{0}]_{k}  + [\boldw_{0}]_{|U| + k'} + [\boldM]_{k,k'}.
\end{align*}

\begin{lemm}
\cite{jaggi2010simple} For any non-zero matrix $\boldM \in \mathbbR^{d\times n}$ and $\eta$:
\[
\|\boldM\|_{\textnormal{tr}} \leq \frac{\eta}{2},
\]
iff $\exists$ symmetric matrices $\boldG \in \mathbbR^{d\times d}$ and $\boldH \in \mathbbR^{n \times n}$ 
\begin{align*}
\boldW = \left[
\begin{array}{cc}
\boldG & \boldM \\
\boldM^\top & \boldH\\
\end{array}
\right]\succeq 0,
~\textnormal{tr}(\boldG) + \textnormal{tr}(\boldH) = \eta.
\end{align*}
\end{lemm}

Based on Lemma~1, the optimization problem Eq.~\eqref{eq:CFM-optimization} is equivalent to
\begin{align}
\label{eq:CFM-optimization3}
\min_{\boldw,\boldM} \widetilde{J}(\boldw, \boldM) + \lambda_1 \|\boldw\|_2^2 \hspace{0.3cm} \text{s.t.}~ \|\boldM \|_{\text{tr}} \leq \frac{\eta}{2},
\end{align}
where 
\begin{align*} 
\widetilde{J}(\boldw,\boldM) := \sum_{(k,k') \in \Omega} ([{\boldA}]_{k,k'} - w_0 - [\boldw_{0}]_{k}  - [\boldw_{0}]_{|U| + k'} - [\boldM]_{k,k'})^2,
\end{align*}
and $\Omega$ is the set of observed values in $\boldA$. If we set $\boldw = \boldzero$, the optimization problem is equivalent to matrix factorization with nuclear norm regularization \cite{jaggi2010simple}; CFM includes convex matrix factorization as a special case. Since we would like to have a low-rank matrix $\boldM$ of the user-item matrix $\boldA$ for recommendation, Eq.~\eqref{eq:CFM-optimization} is a natural formulation for convex FMs. Note that, even though CFM resembles the matrix factorization \cite{jaggi2010simple}. the MF method cannot incorporate side information, while CFM can deal with side-information by concatenating it to vector $\boldx$. That is, intrinsically, the MF method \cite{jaggi2010simple} and CFM are different.

\subsection{Hazan's Algorithm}
For optimizing $\boldW$, we adopt Hazan's algorithm \cite{hazan2008sparse}. It only needs to compute a leading eigenvector of a sparse $d\times d$ matrix in each iteration, and thus it scales well to large problems. Moreover, the proposed CFM update formula is extremely simple, and hence useful for practitioners. 
 The Hazan's algorithm for CFM is summarized in Algorithm \ref{alg:alg}.

\begin{algorithm}[t]
\caption{ CFM with Hazan's Algorithm}
\label{alg:alg}
\begin{algorithmic}
\STATE Rescale loss function $J_{\eta}(\boldW) = \|\boldR(\boldy -\boldf_Q(\boldX;\eta\boldW)\|_2^2$.
\STATE Initialize $\boldW^{(1)}$.
\FORALL {$t=0,1\ldots,T= \lceil\frac{4C_f}{\epsilon}\rceil$}
\STATE Compute $\boldp^{(t)}=\text{approxEV}\big(-\nabla J_{\eta}(\boldW^{(t)}),\frac{C_f}{(t+1)^2}\big)$
\STATE $\widehat{\alpha}_t:=\frac{2}{t+2}$ (\text{or} $\widehat{\alpha}_t = \argmin_{\alpha}~ J_\eta(\boldW^{(t)}+\alpha (\boldp^{(t)}\boldp^{(t)\top} - \boldW^{(t)}))$)%
\STATE $\boldW^{(t+1)}=\boldW^{(t)}+\widehat{\alpha}_t (\boldp^{(t)}\boldp^{(t)\top} - \boldW^{(t)})$
\ENDFOR
\RETURN $\boldW^{(T)}$.
\end{algorithmic}
\end{algorithm}


\vspace{.1in}
\noindent {\bf Derivative computation:} The objective function $J(\boldW)$ can be equivalently written as
\begin{align*}
J(\boldW) =  \sum_{i=1}^{n}\sum_{j=1}^n {\boldC}_{ij} (y_i - \boldf_Q(\boldx_i;\boldW))(y_j - \boldf_Q(\boldx_j;\boldW)).
\end{align*}
Then, $\nabla J(\boldW^{(t)})$ is given as
\begin{align*}
\nabla J(\boldW^{(t)}) 
&=  \boldX \boldD^{(t)} \boldX^\top, \\
\boldD^{(t)} &= -\text{diag}(\sum_{j=1}^n {\boldC}_{1j} (y_j - f_Q(\boldx_1;\boldW)), \ldots, \sum_{j=1}^n {\boldC}_{nj} (y_j - f_Q(\boldx_n;\boldW)\Large),
\end{align*}
where we use $\frac{\partial \text{tr}(\boldW \boldx\boldx^\top)}{\partial \boldW} = \boldx \boldx^\top$.  Since the derivative is written as $\boldX\boldD^{(t)} \boldX^\top$,  the eigenvalue decomposition can be obtained without storing $\nabla J(\boldW^{(t)})$ in memory. Moreover, since the matrix $\boldX$ is a sparse matrix, we can efficiently obtain the leading eigenvector by the Lanczos method. We can use a standard eigenvalue decomposition package to compute the approximate eigenvector by the "$\text{approxEV}$" function. For example in Matlab, we can obtain the approximate eigenvector $\boldp$ by the function $[\boldp, l] = \text{eigs}(-\nabla J(\boldW^{(t)}), 1, '\text{LA}',\text{Options.tol=}\frac{C_f}{(t+1)^2}$)\sloppy, where $l$ is the corresponding eigenvalue.

The proposed CFM optimization requires a matrix inversion (i.e., $O(n^3)$) for computing $\boldC$ in $\boldD^{(t)}$, and it is not feasible if the dimensionality $d$ is large. For example in user-item recommendation task, the total dimensionality of the input can be \emph{the number of users} + \emph{the number of items}. In such cases, the dimensionality can be $10^6$ or more. However, fortunately, the input matrix $\boldX$ is extremely sparse, and we can efficiently compute $\boldD^{(t)}$ by using a conjugate gradient method whose time complexity is $O(n)$.

$\boldD^{(t)}$ can be written as
\begin{align*}
\boldD^{(t)} &= \text{diag}(\boldC \bar{\boldy}^{(t)}) =\text{diag}((\boldR^\top \boldR + \lambda_1 \boldH^\top \boldH)\bar{\boldy}^{(t)}),
\end{align*}
where $\bar{\boldy}^{(t)} = \boldy - \boldf_Q(\boldX;\boldW^{(t)})$. Since the number of samples $n$ tends to be larger than the dimensionality $d$ in factorization machine settings, $\boldZ\boldZ^\top$ becomes full rank. Namely, we can safely make the regularization parameter $\lambda_1 =0$. In such case,  $\boldD^{(t)}$ is given as
\begin{align*}
\boldD^{(t)} 
&= \text{diag}(\bar{\boldy}^{(t)} - \boldZ^\top \widehat{\boldw}^{(t)}),
\end{align*}
where we use $\boldC  = \boldR^\top \boldR = \boldI - \boldZ^\top (\boldZ \boldZ^\top)^{-1}\boldZ$.  The $\widehat{\boldw}^{(t)} = (\boldZ \boldZ^\top)^{-1}\boldZ\bar{\boldy}^{(t)}$ is obtained by solving
\begin{align}
\label{eq:linear-term}
\boldZ^\top \boldw = \bar{\boldy}^{(t)},
\end{align}
where $\boldw^{(t)}$ can be efficiently obtained by a conjugate gradient method with time complexity $O(n)$. Thus, we can compute $\boldD^{(t)}$ without computing the matrix inverse $(\boldZ\boldZ^\top)^{-1}$. To further speed up conjugate gradient method, we use a preconditioner and the previous solution $\boldw^{(t-1)}$ as the initial solution.

 Finally, we compute $\boldD^{(t)}$ as
\begin{align*}
{\boldD}^{(t)} 
&= \text{diag}(\boldy -  \boldf(\boldX;\widehat{\boldw}^{(t)},\widehat{\boldW}^{(t)}) ).
\end{align*}
The diagonal elements of $\boldD^{(t)}$ are the differences between the observed outputs and the model predictions at the $t$-th iteration. Note that, in our CFM optimization, we eliminate $\boldw$ and only optimize for $\boldW$; however, the $\boldw$ is implicitly estimated in Hazan's algorithm. 

\vspace{.1in}
\noindent {\bf Complexity:} Iteration $t$ in Algorithm~\ref{alg:alg} includes computing an approximate leading eigenvector of a sparse matrix with $n$ non-zero elements and an estimation of $\boldw$, which require $O(n)$ computation using Lanczos algorithm and $O(n)$ computaiton using conjugate gradient descent, respectively. Thus, the entire computational complexity of the proposed method is $O(Tn)$, where $T$ is the total number of iterations in Hazan's algorithm.

\vspace{.1in}
\noindent{\bf Optimal step size estimation:} Hazan's algorithm assures $\boldW$ converges to a global optimum with using the step size $\alpha_t = 2/(2+t), t = 0,1,\ldots,T$ \cite{jaggi2010simple}. However, this is in practice slow to converge. Instead, we choose the $\alpha_k$ that maximally decreases the objective function $J(\boldW)$. The optimal $\alpha_k$ can be obtained by solving the following equation:
\begin{align*}
\widehat{\alpha}_t &\!=\! \argmin_{\alpha}~ J(\boldW^{(t+1)}) \\
&\!=\! \argmin_{\alpha} \left\|\boldR \!\left(\boldy \!-\! \boldf_Q(\boldX; (1\!-\!\alpha)\boldW^{(t)} \!+\! \alpha \boldu^{(t)} {\boldu^{(t)}}^\top)\!\right)\!\right\|_2^2.
\end{align*}
Taking the derivative with respect to $\alpha$ and solving the problem for $\alpha$, we have
\begin{align}
\label{eq:optimal_step}
\widehat{\alpha}_t = \frac{(\boldy -\boldf_Q(\boldX;\boldW^{(t)}))^\top \boldR (\boldf_Q(\boldX; \boldp^{(t)} {\boldp^{(t)}}^\top - \boldW^{(t)})) }{\|\boldR (\boldf_Q(\boldX; \boldp^{(t)} {\boldp^{(t)}}^\top - \boldW^{(t)}) \|_2^2}.
\end{align}
The  computation of $\alpha_t$ involves the matrix inversion of $\boldR$. However, by using the same technique as in the derivative computation, we can efficiently compute $\alpha_t$.

\vspace{.1in}
\noindent{\bf Update $\boldW$:} When the input dimension $d$ is large, storing the feature-feature interaction matrix $\boldW$ is not possible. To avoid the memory problem,  we update $\boldW^{(t)}$ as
\begin{align*}
\boldW^{(t+1)} &= \boldP^{(t+1)} \text{diag}(\boldlambda^{(t+1)}) {\boldP^{(t+1)}}^\top,\\
\boldP^{(t+1)} &= [\boldP^{(t)}~\boldp^{(t)}] \in \mathbbR^{d \times (t+1)},\\
\lambda^{(t+1)}_k &=\left\{ \begin{array}{ll}
(1-\widehat{\alpha}_t)\lambda^{(t)}& (k<t) \\
\widehat{\alpha}_t & (k = t+1) \\
\end{array} \right.,
\end{align*}
where $\boldlambda^{(t+1)} \in \mathbbR^{t+1}$. Thus, we only need to store $\boldP^{(t+1)} \in \mathbbR^{d \times (t+1)}$ and $\boldlambda^{(t+1)}$ at the $(t+1)$-th iteration. In practice, Hazan's algorithm converges with $t=100$ (see experiment section), so the required memory for Hazan's algorithm is reasonable. 

\vspace{.1in}
\noindent{\bf Prediction:} Let us define $\boldU = \boldP \text{diag}(\boldlambda)^{1/2} \in \mathbbR^{d \times t}$ such that  $\boldW = \boldU\boldU^{\top}$. 
Then, we can efficiently compute the output as 
\begin{align*}
f(\boldx; \boldw,{\boldW})= {\boldw}^\top \boldz + \frac{1}{2}\left(\|{\boldU}^\top\boldx\|_2^2  -  (\boldx \circ \boldx)^\top ({\boldU}\circ {\boldU})\boldone\right).
\end{align*}
The time complexities of the terms are $O(d)$, $O(d(t+1))$, and $O(d)$, respectively.

\subsection{Tensor completion with CFM}
In this section, we formulate a Tucker based tensor completion problem \cite{tomioka2010estimation,tomioka2013convex} as a CFM problem.

Let us denote the input 3-way tensor as $\calY \in \mathbbR^{n_1 \times n_2 \times n_3}$, where $n_1$, $n_2$, and $n_3$ are the number of samples in each mode. In this paper, we consider the following regularization based learning model:
\begin{align}
\label{eq:Tucker}
\min_{\{\calM^{(m)}\}_{m = 0}^3} \hspace{0.3cm} \sum_{(i,j,k) \in \Omega} \left([\calY]_{i,j,k} - [\calM^{(0)}]_{i,j,k} - \sum_{m = 1}^{3} [\calM^{(m)}]_{i,j,k}\right)^2 + \lambda \sum_{m = 1}^{3} \|\boldM^{(m)}_{(m)}\|_{\text{tr}},
\end{align}
where $\calM^{(0)} \in \mathbbR^{n_1 \times n_2 \times  n_3}$ is the bias tensor, $\calM^{(m)} \in \mathbbR^{n_1 \times n_2 \times n_3}$ is the $m$-th mode tensor, $\lambda \geq 0$ is the regularization parameter, $\boldM^{(m)}_{(m)}$ is the unfolded matrix with respect to the $m$-th mode, $\boldM^{(1)}_{(1)} \in \mathbbR^{n_1 \times n_2n_3}$, $\boldM^{(2)}_{(2)} \in \mathbbR^{n_2 \times n_1n_3}$, and $\boldM^{(3)}_{(3)} \in \mathbbR^{n_3 \times n_1n_2}$.  The final goal of this paper is to learn $\calM$ from $\calY$ by minimizing $J(\calM)$. Now, we reformulate Eq.~\eqref{eq:Tucker} by CFM. 

Let us define the pooled matrix:
\begin{align*}
\boldW = \left[
\begin{array}{ccccccc}
\boldG_1 & \boldM_{(1)}^{(1)} & &  & &&\\
{\boldM_{(1)}^{(1)}}^\top & \boldH_1 &  &&&& \\
&  & & \boldG_2 &  \boldM_{(2)}^{(2)}& &\\
&  & & {\boldM_{(2)}^{(2)}}^\top &  \boldH_2 & &\\
&  & &  &  &\boldG_3& {\boldM_{(3)}^{(3)}} \\
&  & &  &  & {\boldM_{(3)}^{(3)}}^\top & \boldH_3\\
\end{array}
\right] \in \mathbbR^{d \times d}, \boldW \succeq 0,
\end{align*}
where $d = \sum_{m = 1}^3 n_m + \sum_{m = 1}^3 \sum_{m' = m +1}^3 n_m n_{m'}$. Note, the off-diagonal matrices are not important for deriving optimization algorithm, and thus, we omit them here. Moreover, since the matrix $\boldW$ is a positive semi-definite matrix, we can decompose it as
\begin{align*}
\boldW = \left[
\begin{array}{c}
\boldU_{1}  \\
\boldV_1 \\
\vdots \\
\boldU_3 \\
\boldV_3 \\
\end{array}
\right]
\left[
\begin{array}{ccccc}
\boldU_{1}^\top & \boldV_1^\top & \ldots &\boldU_3^\top &\boldV_3^\top \\
\end{array}
\right].
\end{align*}

\begin{lemm} \cite{choi2014dfacto}
For a 3-way tensor case, we have:
\begin{align*}
[\calM^{(1)}]_{i,j,k} &= [\boldM^{(1)}_{(1)}]_{i,n_2(k-1) + j}, \\
[\calM^{(2)}]_{i,j,k} &= [\boldM^{(2)}_{(2)}]_{j,n_3(i-1) + k}, \\
[\calM^{(3)}]_{i,j,k} &= [\boldM^{(3)}_{(3)}]_{k,n_1(j-1) + i}.
\end{align*}
\end{lemm}

Then, we can rewrite $\sum_{m = 1}^{3} [\calM^{(m)}]_{i,j,k}$ as
\begin{align*}
\sum_{m = 1}^{3} [\calM^{(m)}]_{i,j,k} &= \frac{1}{2}\text{tr}\left(\boldW (\boldx_{i,j,k} \boldx_{i,j,k}^\top - \text{diag}(\boldx_{i,j,k} \circ \boldx_{i,j,k})) \right), \\
\boldx_{i,j,k}^{(1)} &=[\overbrace{0~\cdots 0~\underbrace{1}_{i}~0~\cdots 0}^{n_1}~\overbrace{0 ~\cdots 0~\underbrace{1}_{n_2(k-1) + j}~0 \cdots 0}^{n_2n_3}]^\top \in \mathbbR^{n_1 + n_2n_3}, \\
\boldx_{i,j,k}^{(2)} &=[\overbrace{0~\cdots 0~\underbrace{1}_{j}~0~\cdots 0}^{n_2}~\overbrace{0 ~\cdots 0~\underbrace{1}_{n_3(i-1) + k}~0 \cdots 0}^{n_1n_3}]^\top \in \mathbbR^{n_2 + n_1n_3}, \\
\boldx_{i,j,k}^{(3)} &=[\overbrace{0~\cdots 0~\underbrace{1}_{k}~0~\cdots 0}^{n_3}~\overbrace{0 ~\cdots 0~\underbrace{1}_{n_1(j-1) + i}~0 \cdots 0}^{n_1n_2}]^\top \in \mathbbR^{n_3 + n_1n_2}, \\
\boldx_{i,j,k} &= [{\boldx_{i,j,k}^{(1)}}^\top~{\boldx_{i,j,k}^{(2)}}^\top ~{\boldx_{i,j,k}^{(3)}}^\top]^\top \in \mathbbR^d.
\end{align*}

For the bias tensor $\calM^{(0)}$, we parametrize it as
\begin{eqnarray}
[\calM^{(0)}]_{i,j,k} = \boldw^\top \boldx_{i,j,k} = \left[
\begin{array}{l}
\boldw_1\\
\boldzero_{n_2n_3} \\
\boldw_2 \\
\boldzero_{n_1n_3} \\
\boldw_3 \\
\boldzero_{n_1n_2}
\end{array}
\right]^\top \boldx_{i,j,k},
\end{eqnarray}
where $\boldw \in \mathbbR^d$, $\boldw_1 \in \mathbbR^{n_1}$, $\boldw_2 \in \mathbbR^{n_2}$, and $\boldw_3 \in \mathbbR^{n_3}$. Note, we use this parameterization, since the number of dimension $d$ can be much bigger than the number of non-zero elements $n$ and it is hard to solve Eq.~\eqref{eq:linear-term}. 

\begin{lemm}
\label{lemm1}
For the matrices $\boldM_{(1)}^{(1)} \in \mathbbR^{n_1 \times n_2 n_3}, \boldM_{(2)}^{(2)} \in \mathbbR^{n_2 \times n_1 n_3}, \boldM_{(3)}^{(3)} \in \mathbbR^{n_3 \times n_1 n_2}$ and $\eta$:
\[
\sum_{m = 1}^3 \|\boldM_{(m)}^{(m)}\|_{\textnormal{tr}}
 \leq \frac{\eta}{2}
 \]
 iff
 \begin{align*}
 \boldW = \left[
\begin{array}{ccccccc}
\boldG_1 & \boldM_{(1)}^{(1)} & &  & &&\\
{\boldM_{(1)}^{(1)}}^\top & \boldH_1 &  &&&& \\
&  & & \boldG_2 &  \boldM_{(2)}^{(2)}& &\\
&  & & {\boldM_{(2)}^{(2)}}^\top &  \boldH_2 & &\\
&  & &  &  &\boldG_3& {\boldM_{(3)}^{(3)}} \\
&  & &  &  & {\boldM_{(3)}^{(3)}}^\top & \boldH_3\\
\end{array} 
\right]  \succeq 0 ,
\end{align*}
and $\sum_{m = 1}^{3} \textnormal{tr}(\boldG_m) + \textnormal{tr}(\boldH_m) = \eta$.

\noindent Proof: This is a variation of the Lemma1 of \cite{jaggi2010simple}.  From the characterization:
\[
\sum_{m = 1}^3 \|\boldM_{(m)}^{(m)}\|_{\textnormal{tr}} = \min_{\{\boldU_m \boldV_m^\top = \boldM_{(m)}^{(m)}\}_{m =1}^3} \hspace{0.3cm} \frac{1}{2}\sum_{m = 1}^3 \left(\|\boldU_m\|_{F} + \|\boldV_m\|_F\right)
\]
we have that $\exists ~\boldU_m, \boldV_m, \boldU_m \boldV_m^\top = \boldM_{(m)}^{(m)}, m = 1,2,3$ s.t. 
\[
2\sum_{m=1}^3 \|\boldM_{(m)}^{(m)}\|_{\textnormal{tr}} = \sum_{m =1}^3 \|\boldU_m\|_F^2 + \|\boldV_m\|_F^2 = \sum_{m =1}^3  \textnormal{tr}(\boldU_m\boldU_m^\top) + \textnormal{tr}(\boldV_m\boldV_m^\top) \leq \eta.
\]
 That is, we have
\begin{align*}
 \boldW = \left[
\begin{array}{ccccccc}
\boldU_1 \boldU_1^\top & \boldM_{(1)}^{(1)} & &  & &&\\
{\boldM_{(1)}^{(1)}}^\top & \boldV_1 \boldV_1^\top &  &&&& \\
&  & & \boldU_2 \boldU_2^\top &  \boldM_{(2)}^{(2)}& &\\
&  & & {\boldM_{(2)}^{(2)}}^\top &  \boldV_2 \boldV_2^\top & &\\
&  & &  &  &\boldU_3 \boldU_3^\top& {\boldM_{(3)}^{(3)}} \\
&  & &  &  & {\boldM_{(3)}^{(3)}}^\top & \boldV_3 \boldV_3^\top\\
\end{array}
\right],
\end{align*}
where $\textnormal{tr}(\boldW) \leq \eta$ and $\boldW \succeq 0$. If $s = \textnormal{tr}(\boldW) < \eta$, we can add $(t -s) \bolde_1\bolde_1^\top$ to $\boldU_1 \boldU_1^\top$, and we have $\textnormal{tr}(\boldW) = \eta$.

If the matrix is symmetric and positive semi-definite, we can decompose $\boldW$ as
\begin{align*}
\boldW = \left[
\begin{array}{c}
\boldU_{1}  \\
\boldV_1 \\
\vdots \\
\boldU_3 \\
\boldV_3 \\
\end{array}
\right]
\left[
\begin{array}{ccccc}
\boldU_{1}^\top & \boldV_1^\top & \ldots &\boldU_3^\top &\boldV_3^\top \\
\end{array}
\right],
\end{align*}
such that $\boldU_m\boldV_m^\top = \boldM_{(m)}^{(m)}, m = 1, 2, 3$ and $\eta = \sum_{m =1}^3  \textnormal{tr}(\boldU_m\boldU_m^\top) + \textnormal{tr}(\boldV_m\boldV_m^\top) =  \sum_{m =1}^3  \|\boldU_m\|_{F}^2 + \|\boldV_m\|_F^2$. \proofend 
\end{lemm}

Based on the Lemma \ref{lemm1}, we can rewrite the optimization problem as
\begin{align*}
\min_{\boldw, \boldW} &\hspace{0.3cm}  \sum_{(i,j,k) \in \Omega} \left([\calY]_{i,j,k} - \boldw^\top \boldx_{i,j,k} - \text{tr}\left(\boldW (\boldx_{i,j,k} \boldx_{i,j,k}^\top - \text{diag}(\boldx_{i,j,k} \circ \boldx_{i,j,k})) \right)\right)^2 \\
\text{s.t} &\hspace{0.3cm} \boldW \succeq 0, \text{tr}(\boldW) = \eta.
\end{align*}
Since this is a CFM problem, we can efficiently solve it with Hazan's algorithm. 

\section{Related Work}
First of all, the same problem setting as in our work has been addressed quite recently \cite{convex_fm}, being independent of our work. The key difference between the proposed method and \cite{convex_fm} is that our approach is based on a single convex optimization problem for the interaction term $\boldW$. The approach \cite{convex_fm} uses a block-coordinate descent (BCD) algorithm for optimization, optimizing the linear and quadratic terms alternatively. That is, they alternatingly solve the following two update equations until convergence:
\begin{align*}
\widehat{\boldw}^{(t+1)} &= \argmin_{\boldw} \hspace{0.1cm} \|\boldy - \boldf(\boldX;\boldw,\boldW^{(t)})\|_2^2 + \lambda_1 \|\boldw\|_2^2, \\
\widehat{\boldW}^{(t+1)} &= \argmin_{\boldW} \hspace{0.1cm} \|\boldy - \boldf(\boldX;\widehat{\boldw}^{(t+1)},\boldW)\|_2^2 + \lambda_2 \|\boldW\|_{\text{tr}},
\end{align*}
while our proposed approach is simply given as
\begin{align*}
\widehat{\boldW} &= \argmin_{\boldW \succeq 0} \hspace{0.1cm} \|\boldR(\boldy -\boldf_Q(\boldX;\boldW))\|_2^2,\text{s.t.}~\|\boldW\|_{\text{tr}} = \eta.
\end{align*}
Hence, the BCD algorithm needs to iterate the sub-problem for $\boldW$ until convergence for obtaining the globally optimal solution. 

Let us employ an $O(n)$ algorithm for the trace norm minimization in BCD; then the entire complexity is $O(T'Tn)$ where $T'$ is the BCD iteration and $T$ is the iteration of the sub-problem. On the other hand, our algorithm's complexity is $O(Tn)$. Another difference is that our optimization approach includes the matrix factorization with nuclear norm regularization as a special case, while it is unclear whether the same holds for the formulation \cite{convex_fm}. Finally, our CFM approach is very easy to implement; the core part of the proposed algorithm can be written within 20 lines in Matlab.    Note, the BCD based approach is more general than our CFM framework; it can be used for other loss functions such as logistic loss and it does not require the positive definiteness condition for $\boldW$.

The convex variant of matrix factorization has been widely studied in machine learning community \cite{fazel2001rank,candes2009exact,ji2009accelerated,bach2008convex,toh2010accelerated,tomioka2010estimation,tomioka2011statistical,liu2012implementable}. The key idea of the convex approach is to use the trace norm regularizer, and the optimization problem is given as
\begin{align}
\label{eq:MF}
\widehat{\boldM} = \argmin_{\boldM}~ \|\calP_{\Omega}(\boldA) - \calP_{\Omega}(\boldM)\|_{F}^2 + \lambda \|\boldM\|_{\text{tr}},
\end{align}
where $\Omega$ is the set of observed value in $\boldA$, $[\calP_{\Omega}(\boldA)]_{i,j} = [\boldA]_{i,j}$ if $i,j \in \Omega$ and 0 otherwise, and $\|\cdot\|_F$ is the Frobenius norm. Since Eq.\eqref{eq:MF} and Eq.\eqref{eq:CFM-optimization3} are equivalent when $\boldw = \boldzero$, the convex matrix factorization can be regarded as a special case of CFM.  

To optimize Eq. \eqref{eq:MF}, the singular value thresholding (SVT) method has been proposed \cite{mazumder2010spectral,cai2010singular}, where SVT converges faster in $O(\frac{1}{\sqrt{\epsilon}})$ ($\epsilon$ is an approximate error). However, the SVT approach requires to solve the full eigenvalue decomposition, which is computationally expensive for large datasets. To deal with large data, Frank-Wolfe based approaches have been  proposed including Hazan's algorithm \cite{jaggi2010simple}, corrective refitting \cite{shalev2011large}, and active subspace selection \cite{hsieh2014nuclear}. However, these approaches cannot incorporate user and item bias. Furthermore, it is not straightforward to incorporate side information to deal with \emph{cold start} problems (i.e., recommending an item to a user who has no click information).

To handle \emph{cold start} problems, collective matrix factorization (collective MF) has been proposed \cite{singh2008relational}. The key idea of collective MF is to incorporate side information into matrix factorization. More specifically, we prepare a user $\times$ user meta matrix (e.g., gender, age, etc.) and an item $\times$ item meta matrix  (item category, item title, etc) in addition to a user-item matrix. Then, we factorize all the matrices together. A convex variant of CMF called convex collective matrix factorization (CCMF) has been proposed \cite{bouchard2013convex}. CCMF employs the convex collective norm, which is a generalization of the trace norm to several matrices. Recently, Hazan's algorithm was introduced to CCMF \cite{gunasekarconsistent}. More importantly, it has been theoretically justified that CCMF can give better performance in cold start settings. Since FMs can incorporate side information,  FMs and CCMF are closely related. Actually, CFM can utilize side information and can learn the user and item bias term together; it can be regarded as a generalized variant of CCMF.

\begin{figure}[t!]
\begin{center}
\begin{minipage}[t]{1\linewidth}
\centering
  {\includegraphics[width=0.5\textwidth]{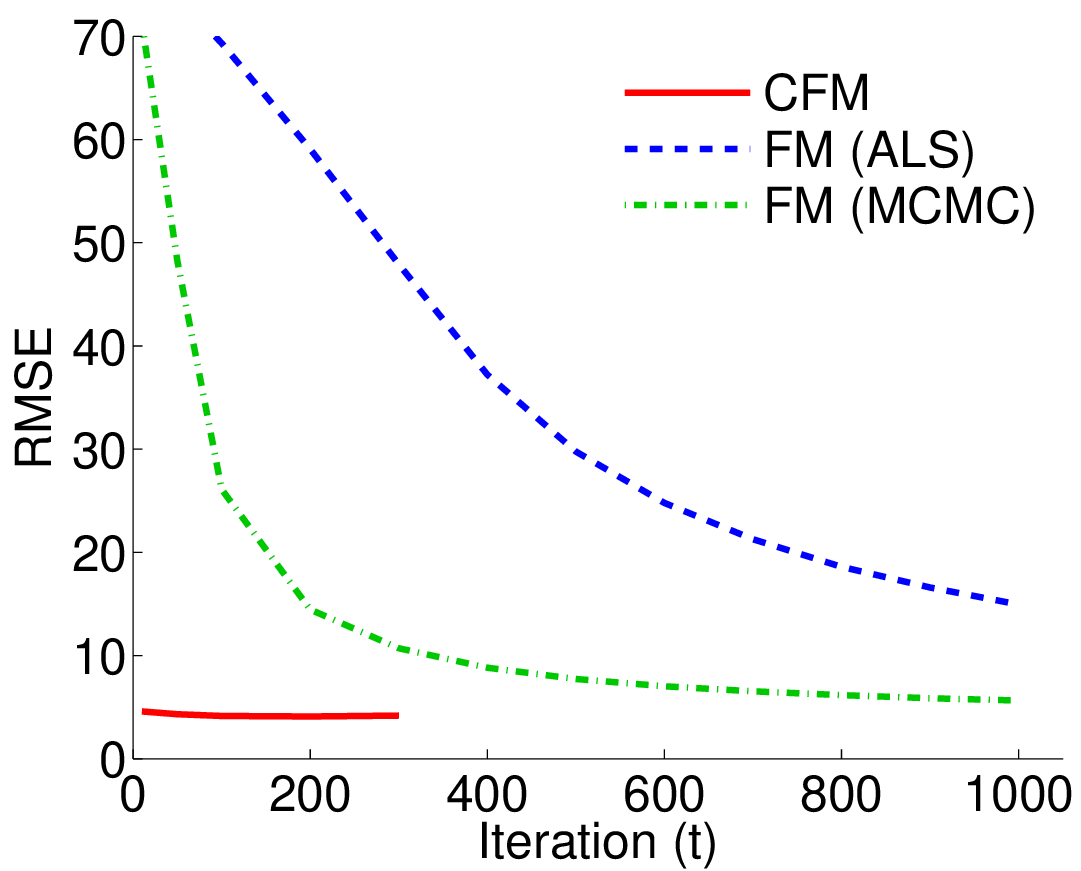}} \\ \vspace{-0.10cm}
\end{minipage}
 \caption{Convergence of the methods: test RMSE of  the synthetic experiment. CFM is the proposed convex factorization machine, FM (ALS) is the factorization machine with ALS optimization, and FM (MCMC) is the factorization machine with MCMC optimization. } 
    \label{fig:result_synth}
\end{center}
\vspace{-.1in}
\end{figure}

\section{Experiments}
We evaluate the proposed method on one synthetic dataset, Movielens data (single matrix), and toxicogenomics data (two-view tensor).

In this paper, we compare CFM with ridge regression, FM (SGD), FM (MCMC) and FM (ALS), where FM (MCMC) is a state-of-the-art FM optimization method. The ridge regression corresponds to the factorization machine with only the linear term $f(\boldx) = w_0 + \boldw_0^\top \boldx$, which is also a strong baseline. To estimate FM models, we use the publicly available libFM package\footnote{\url{http://www.libfm.org}}. For all experiments, the number of latent dimensions of FMs is set to 20, which performs well in practice. For FM (ALS), we experimentally set the regularization parameters as $\lambda_1 = 0$ and $\lambda_2 = 0.01$.  The initial matrices $\boldW$ (for CFM) and $\boldG$ (for FMs) are randomly set (this is the default setting of the libFM package). For CFM, we implemented the algorithm with Matlab. We experimentally set $C_f = 1$, and it works for our experiments.   For all experiments, we use a server with  16 core 1.6GHz CPU and 24G memory. 

When evaluating the performance of CFM and FMs,  we use the root mean squared error (RMSE):
\[
\sqrt{\frac{1}{n_{\text{test}}} \sum_{i = 1}^{n_{\text{test}}} (y^{\ast}_i - \widehat{y}_i)^2},
\]
where $y^{\ast}$ and $\widehat{y}$ are the true and estimated target values, respectively. 

\subsection{Synthetic Experiments}
First, we illustrate how the proposed CFM behaves using a synthetic dataset. 

In this experiment, we randomly generate input vectors $\boldx \in \mathbbR^{100}$ as $\boldx \sim \calN(\boldzero,\boldI)$, and output values as
\begin{align*}
 y = \widetilde{w}_0 + \widetilde{\boldw}^\top \boldx + \sum_{\ell = 1}^d \sum_{\ell' = \ell + 1}^d [\widetilde{\boldW}]_{\ell,\ell'} x_\ell x_{\ell'},
\end{align*}
where
\begin{align*}
\widetilde{w}_0 &\sim \calN(0,1), ~~\widetilde{\boldw}  \sim \calN(\boldzero, \boldI),~~\widetilde{\boldW}_{\ell,\ell'} \sim \text{Uniform}([0~ 1]).
\end{align*}

We use $900$ samples for training and $100$ samples for testing. We run the experiments $5$ times with randomly selecting training and test samples and report the average RMSE scores.  Figure \ref{fig:result_synth} shows the test RMSE for CFM and FMs. As can be seen, the proposed CFM gets the lowest RMSE values with a small number of iterations, while FMs needs many iterations to obtain reasonable performance.  
 
 \begin{figure*}[t!]
\begin{center}
\begin{minipage}[t]{0.45\linewidth}
\centering
  {\includegraphics[width=0.99\textwidth]{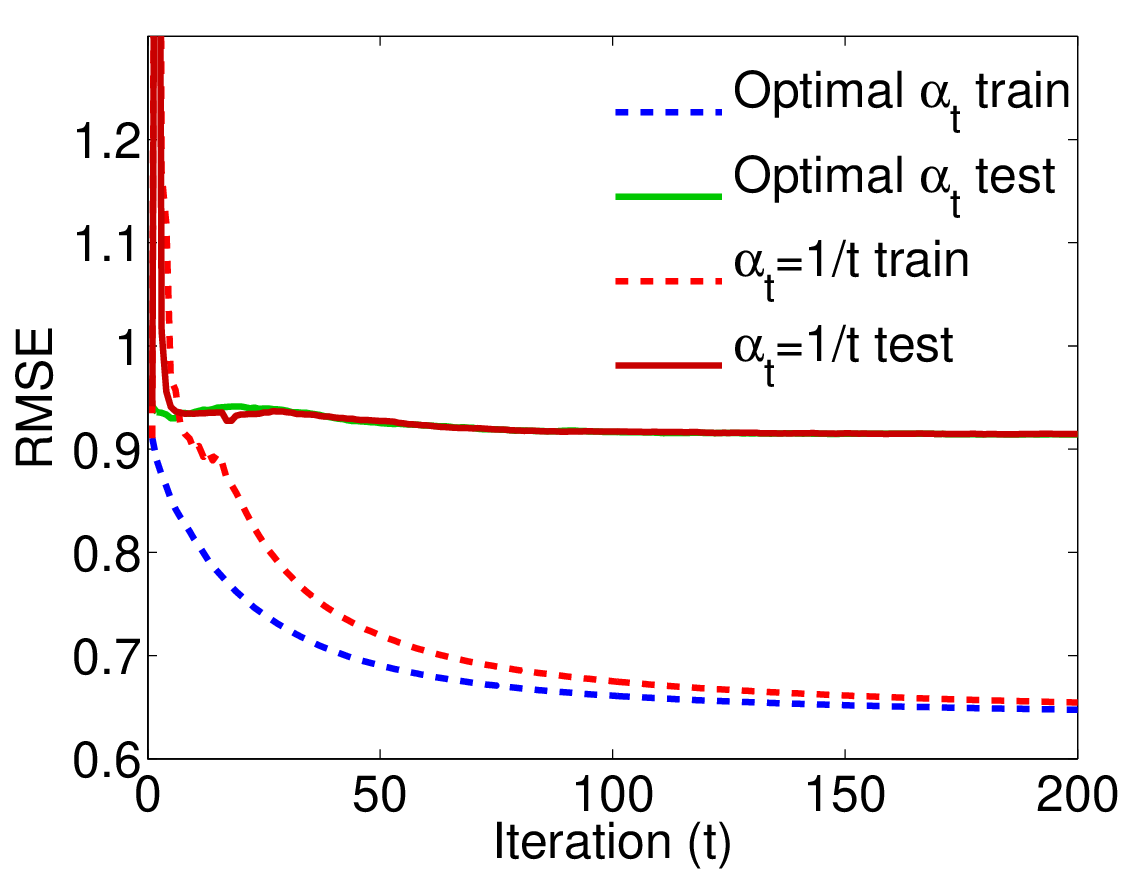}} \\ \vspace{-0.10cm}
(a) Movielens 100K data.
\end{minipage}
\begin{minipage}[t]{0.45\linewidth}
\centering
  {\includegraphics[width=0.99\textwidth]{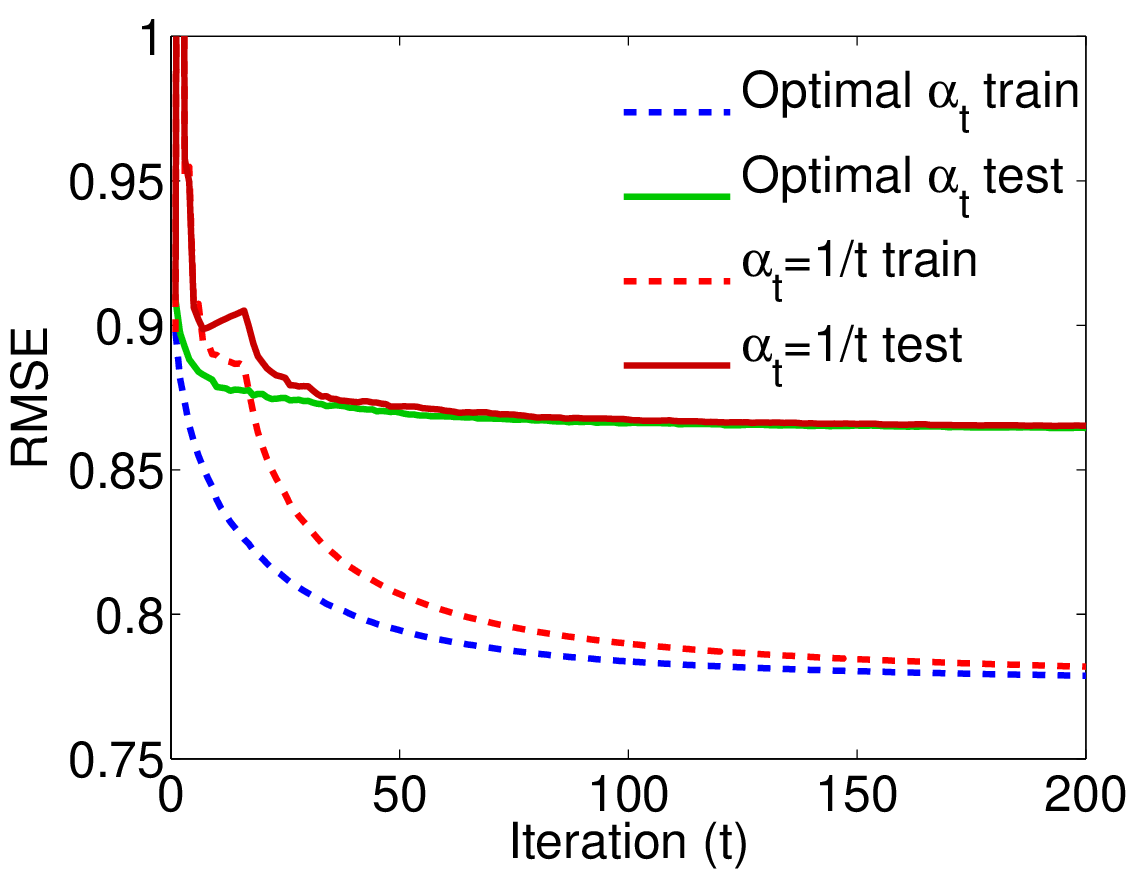}} \\ \vspace{-0.10cm}
(b) Movielens 1M data.
\end{minipage}
\begin{minipage}[t]{0.45\linewidth}
\centering
  {\includegraphics[width=0.99\textwidth]{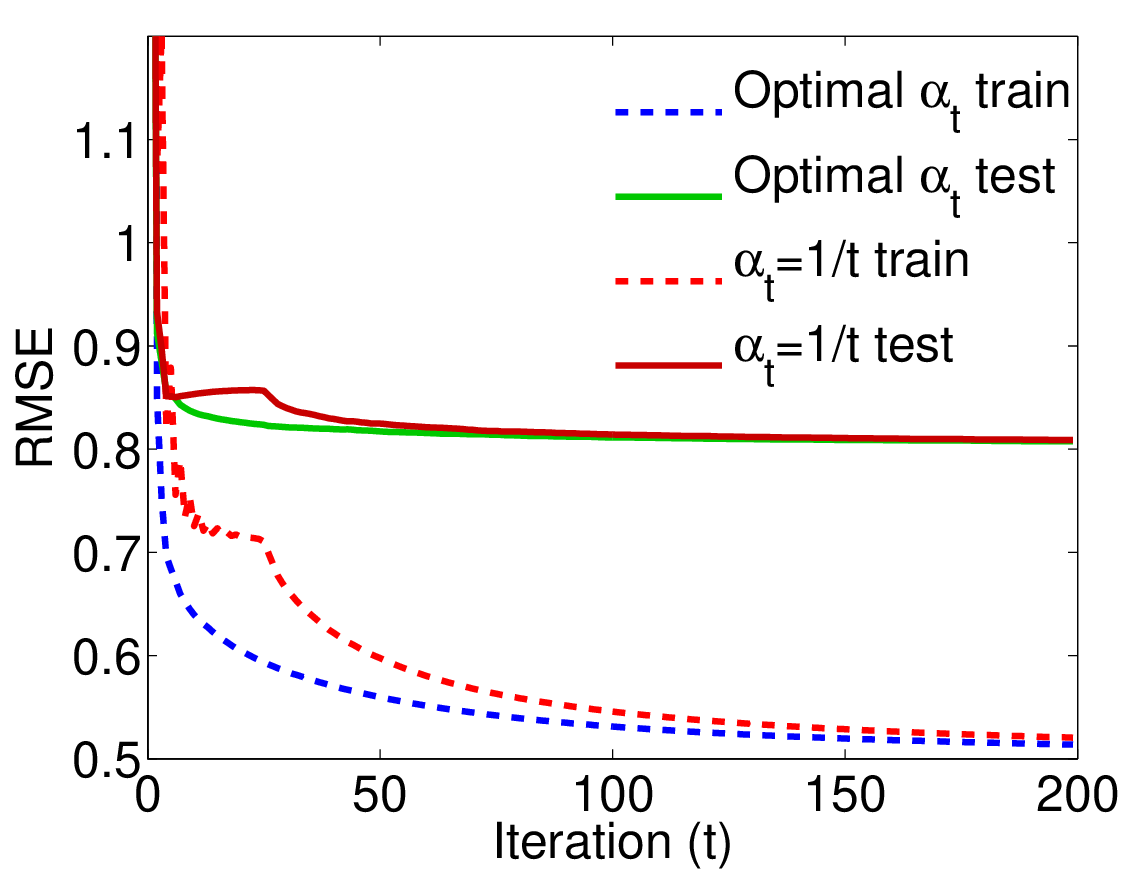}} \\ \vspace{-0.10cm}
(c) Movielens 10M data.
\end{minipage}
\begin{minipage}[t]{0.45\linewidth}
\centering
  {\includegraphics[width=0.99\textwidth]{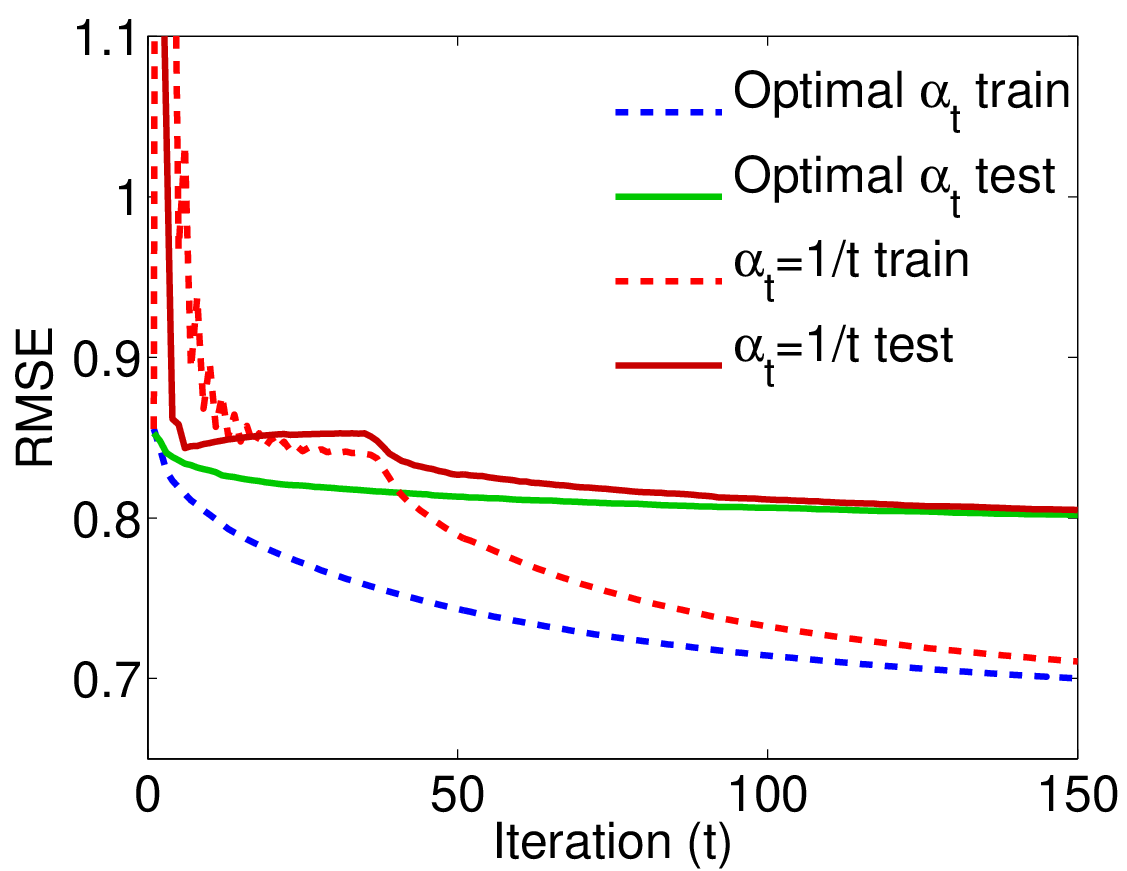}} \\ \vspace{-0.10cm}
(d) Movielens 20M data.
\end{minipage}
 \caption{RMSE over iteration (t). $\alpha_t = \frac{1}{t}$ train and $\alpha_t=\frac{1}{t}$ test are training and test RMSE with using $\frac{2}{2+t}$ stepsize. Optimal $\alpha_t$ train and $\alpha_t$ test are training and test RMSE with using the optimal stepsize Eq.~\eqref{eq:optimal_step}.}
    \label{fig:movie_lens_rmse}
\end{center}
\vspace{-.15in}
\end{figure*}

\begin{figure*}[t!]
\begin{center}
\begin{minipage}[t]{0.45\linewidth}
\centering
  {\includegraphics[width=0.99\textwidth]{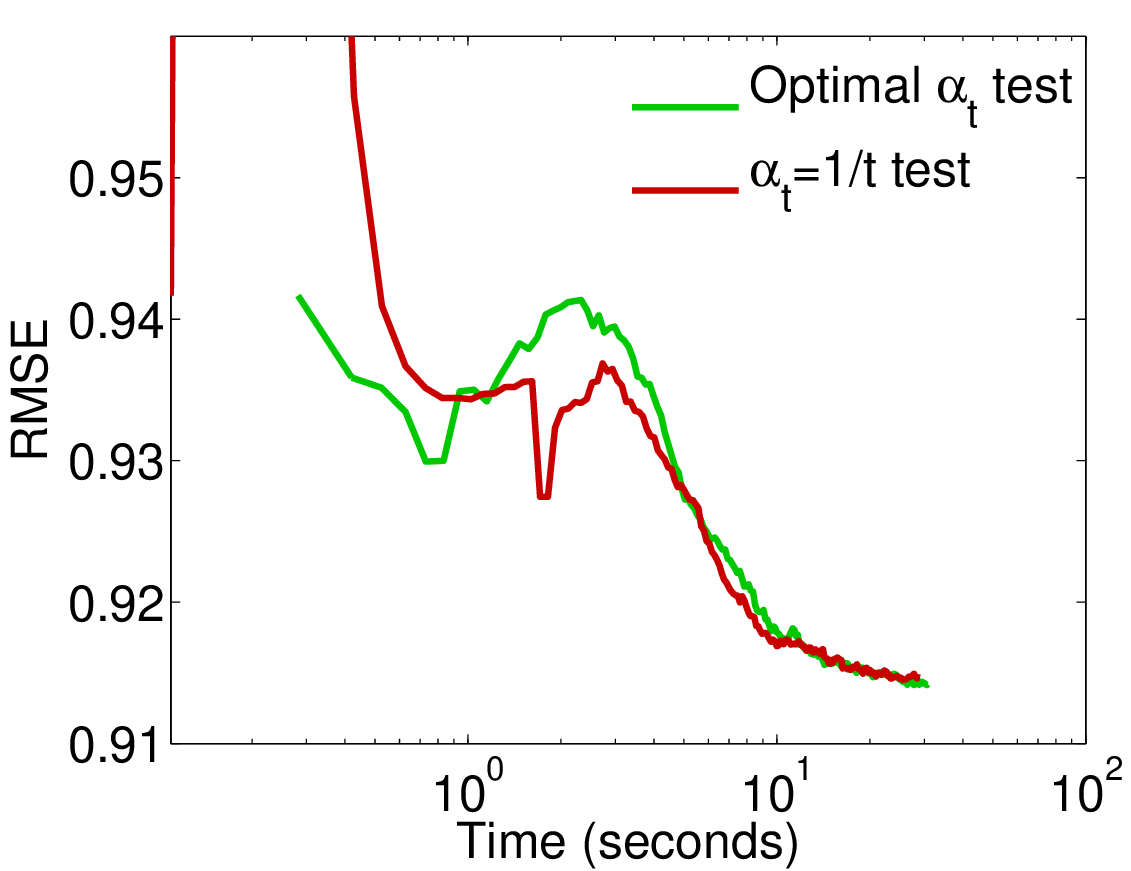}} \\ \vspace{-0.10cm}
(a) Movielens 100K data.
\end{minipage}
\begin{minipage}[t]{0.45\linewidth}
\centering
  {\includegraphics[width=0.99\textwidth]{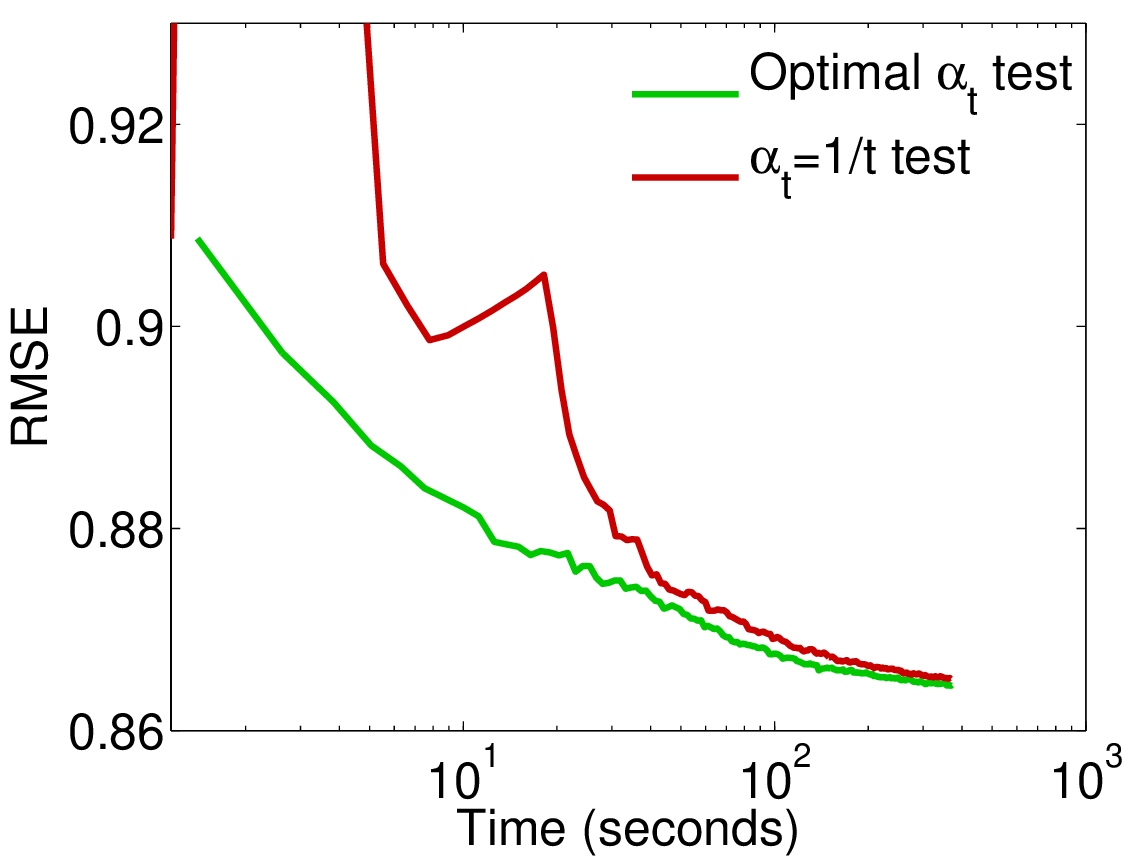}} \\ \vspace{-0.10cm}
(b) Movielens 1M data.
\end{minipage}
\begin{minipage}[t]{0.45\linewidth}
\centering
  {\includegraphics[width=0.99\textwidth]{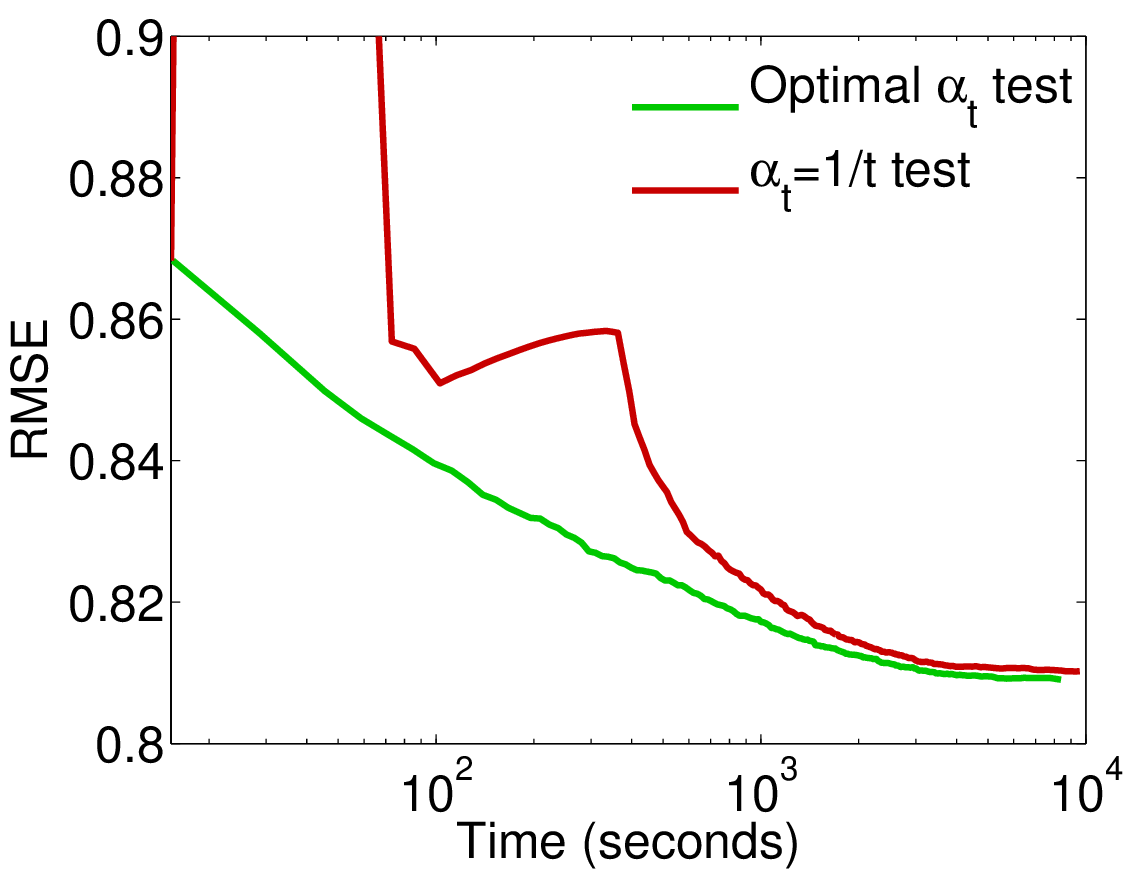}} \\ \vspace{-0.10cm}
(c) Movielens 10M data.
\end{minipage}
\begin{minipage}[t]{0.45\linewidth}
\centering
  {\includegraphics[width=0.99\textwidth]{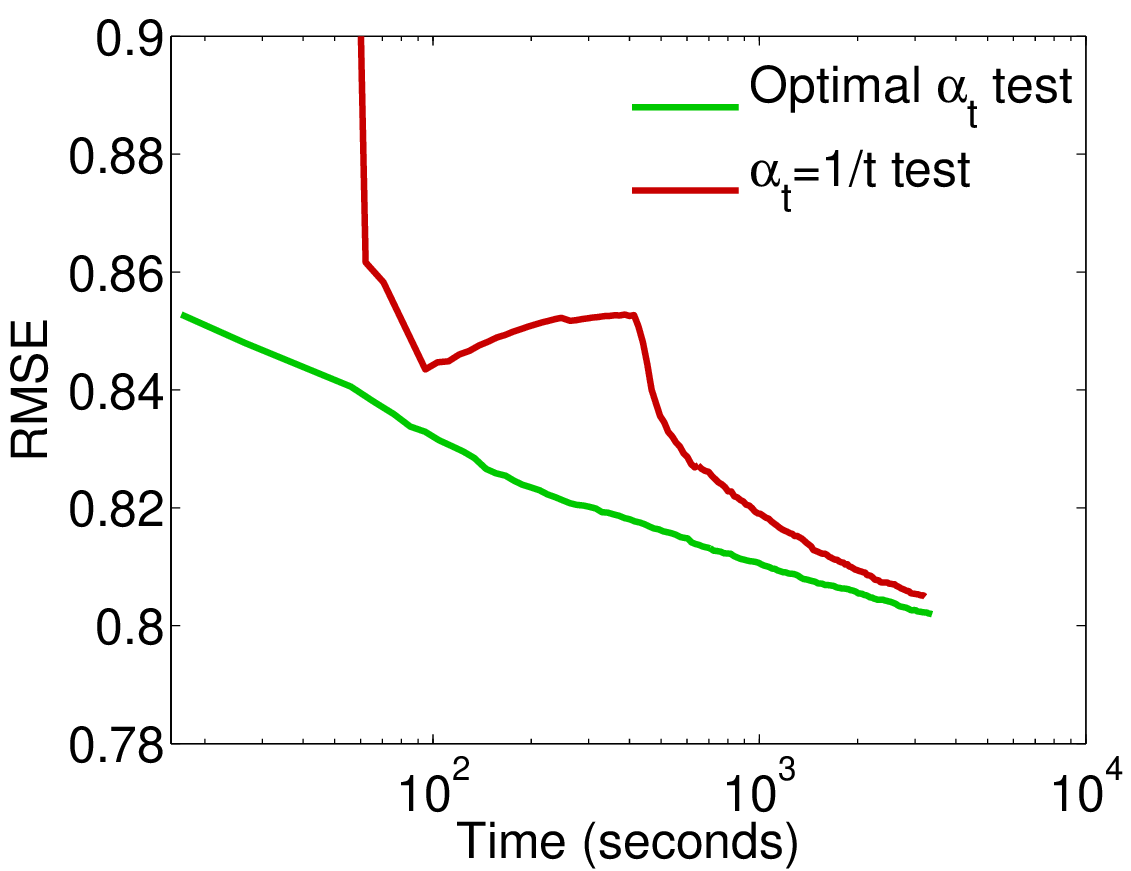}} \\ \vspace{-0.10cm}
(d) Movielens 20M data.
\end{minipage}
 \caption{RMSE over time (seconds). $\alpha_t=\frac{1}{t}$ test is the test RMSE with using $\frac{2}{2+t}$ stepsize. Optimal $\alpha_t$ test is the test RMSE with using optimal stepsize.}
    \label{fig:movie_lens_time}
\end{center}
\vspace{-0.15in}
\end{figure*}
 
\subsection{Recommendation Experiments}
Next, we evaluate our proposed method on the Movielens 100K, 1M, 10M, and 20M datasets \cite{miller2003movielens} (Table \ref{tab:movielens} for dataset details).  In these experiments, we randomly split the observations into 75\% for training and 25\%  for testing. We run the recommendation experiments on three random splits, which is the same experimental setting as in \cite{convex_fm}, and report the average RMSE score. 

\begin{table}[t]
{\small
\begin{center}
\caption{Movielens datasets.\label{tab:movielens}}
\begin{tabular}{|l|c|c|c|c|}
\hline
Dataset & $|U|$ & $|I|$ & $d$ &$|A|=n$ \\ \hline
Movielens 100K & 943 & 1,682 & 2,625& 100,000\\ \hline
Movielens 1M & 6,040 & 3,900 & 9,940& 1,000,209\\ \hline
Movielens 10M & 82,248 & 10,681& 92,929 &10,000,054\\ \hline
Movielens 20M & 138,493 & 27,278 & 165,771 &20,000,263 \\\hline
\end{tabular}
\end{center}}
\end{table} 

\begin{table*}[t!]
{
\begin{center}
\caption{Test RMSE of CFM, CFM (BCD), FMs, and ridge regression for Movielens data sets.\label{tab:result_movielens}}
\begin{tabular}{|l||c|c|c|c|c|c|c|c|}
\hline
Dataset  & CFM & CFM (BCD)& $\text{FM}_{\text{SGD}}$ &$\text{FM}_{\text{ALS}}$ & $\text{FM}_{\text{MCMC}}^{(0.05)}$ &$\text{FM}_{\text{MCMC}}^{(0.1)}$ & Ridge\\ \hline
 100K & 0.915 & 0.93 & 1.078 &   1.242 & 0.905 & 0.901& 0.936\\ \hline
 1M & 0.866& 0.85 &0.943 & 0.981 & 0.877 & 0.846 &0.899\\ \hline
 10M &  0.810 & 0.82 & 0.827  & 0.873  &  0.831 & 0.778 & 0.855\\ \hline
 20M &  0.802& n/a &  0.821  &   0.852 &  0.803  & 0.768 & 0.850 \\\hline
\end{tabular}
\end{center}}
\end{table*} 

 For CFM, the regularization parameter $\eta$ is experimentally set to $2000$ (for 100K), $4000$ (for 1M), $20000$ (for 10M), and $40000$ (for 20M), respectively. For FMs, the rank is set to $20$, which gives overall good performance. To investigate the effect of the initialization parameter, we initialize FM (MCMC) with two parameters $\text{stdev} = 0.05$ and $\text{stdev} = 0.1$, which are the standard deviation of the random variable for initializing $\boldG$. We also report the RMSE of the CFM method of  \cite{convex_fm} for reference.

Figure \ref{fig:movie_lens_rmse} shows the training and test RMSE with the CFM (optimal step size) and the CFM ($\alpha_t = \frac{2}{2+t}$) for the Movielens datasets. For both methods, the RMSE of training and test is converging with a small number of iterations. Overall, the optimal step size based approach  converges faster than the one based on $\alpha_t = \frac{2}{2+t}$. Figure \ref{fig:movie_lens_time} shows the RMSE over computational time (seconds). For large datasets, the CFM achieves reasonable performance in less than an hour. In Table \ref{tab:result_movielens}, we show the RMSE comparison of the proposed CFM with FMs. As we expected, CFM compares favorably with FM (SGD) and FM (ALS), since FM (SGD) and FM (ALS) can be easily trapped at poor locally optimal solutions. Moreover, our CFM method compares favorably with also the CFM (BCD) \cite{convex_fm}. On the other hand, FM (MCMC) can obtain better performance than CFMs (both our formulation and \cite{convex_fm}) for these datasets if we set an appropriate initialization parameter. This is because MCMC tends to avoid poor locally optimal solution if we run the sampler long enough. That is, since the objective function of FMs is non-convex and it has more flexibility than the convex formulation, it can converge to a better solution than CFM if we initialize FMs well. 

\subsection{Prediction in Toxicogenomics}
Next, we evaluated our proposed method on a toxicogenomics dataset \cite{khan2014bayesian}. The dataset contains three sets of matrices representing gene expression and toxicity responses of a set of drugs. The first set called \emph{Gene Expression}, represents the differential expression of 1106 genes in three different cancer types, to a collection of 78 drugs (i.e., $\boldA_l^{(1)} \in \mathbbR^{1106 \times 78}, l = 1,2,3$). The second set, \emph{Toxicity}, contains three dose-dependent toxicity profiles of the corresponding 78 drugs over the three cancers (i.e., $\boldA_l^{(2)} \in \mathbbR^{3 \times 78}, l = 1,2,3$). The gene expression data of the three cancers (Blood, Breast and Prostate) comes from the Connectivity Map \cite{lamb2006cmap} and were processed to obtain differential expression of treatment vs control. As a result, the expression scores represent positive or negative regulation with respect to the untreated level. The toxicity screening data, from the NCI-60 database \cite{shoemaker2006nci60}, summarizes the toxicity of drug treatments in three variables GI50, LC50, and TGI, representing the 50\% growth inhibition, 50\% lethal concentration, and total growth inhibition levels. The data were conformed to represent dose-dependent toxicity profiles for the doses used in the corresponding gene expression dataset.

\vspace{.1in}
\noindent {\bf Predicting both gene and toxicity matrices:}
We compared our proposed method with existing state-of-the-art methods. In this experiment, we randomly split the observations into 50\% for training ($129,466$ elements) and 50\%  for testing ($129,465$ elements), which is the exactly same datasets used in \cite{khan2014bayesian}.  We run the prediction experiments on 100 random splits \cite{khan2014bayesian}, and report the average \emph{relative MSE} score, which is defined as
\[
\sum_{v = 1}^{V} \frac{\|\boldy^{\ast,v} - \widehat{\boldy}^{v}\|_2^2}{\|\boldy^{\ast,v} - \bar{y}^{\ast,v} \boldone \|_2^2},
\]
where $\boldy^{\ast,v}$ is the target score vector,  $\widehat{\boldy}^{v}$ is the estimated score vector, and $\bar{y}^{\ast,v}$ is the mean of elements in $\boldy^{\ast,v}$, $V$ is the number of views. In this experiment, the number of views is $V=2$. Since the number of elements in view~1 and view~2 are different, the \emph{relative} MSE score is more suitable than the root MSE score. We compare our proposed method with ARDCP \cite{morup2009automatic}, CP \cite{carroll1970analysis}, Group Factor Analysis (GFA) \cite{virtanen2012bayesian}, and Bayesian Multi-view Tensor Factorization (BMTF) \cite{khan2014bayesian}. BMTF is a state-of-the-art multi-view factorization method. 

For CFM, we first concatenate all view matrices as 
\begin{align*}
\boldA = \left[
\begin{array}{cc}
\boldA_1^{(1)} & \boldA_1^{(2)} \\
\boldA_2^{(1)} & \boldA_2^{(2)}\\
\boldA_3^{(1)} & \boldA_3^{(2)} 
\end{array}
\right] \in \mathbbR^{3327 \times 78},
\end{align*}
and use this matrix for learning. The regularization parameter $\eta$ is experimentally set to $1000$. To deal with multi-view data, we form the input and output of CFM as

\begin{align*}
\boldx \!&=\! [\overbrace{0 ~\cdots 0~\underbrace{1}_{i\text{-th gene}}~0 \cdots 0}^{3327} ~\overbrace{0 ~\cdots 0~\underbrace{1}_{j\text{-th drug}}~0 \cdots 0}^{78}~ \overbrace{\underbrace{1}_{\text{1st  view}}~0}^{2}]^\top \in \mathbbR^{3407}, \\
y &= [\boldA]_{i,j}.
\end{align*}

Table \ref{tab:result_tox} shows the average relative MSE of the methods. As can be seen, the proposed method outperforms the state-of-the-art methods. 

\begin{table*}[t!]
{
\begin{center}
\caption{Test relative MSE of both gene expression and toxicogenomics data.\label{tab:result_tox}}
\begin{tabular}{|l||c|c|c|c|c|c|c|c|}
\hline
 & \multicolumn{4}{c|}{Multi-view} & \multicolumn{3}{c|}{Single-view}\\
 & {\bf CFM} & BMTF & GFA & ARDCP& CP & ARDCP & CP \\ \hline
Mean & {\bf 0.4037} & 0.4811 & 0.5223 &   0.8919 & 5.3713 & 0.6438 & 5.0699\\ 
StdError& {\bf 0.0163} & 0.0061 &0.0041 & 0.0027 & 0.0310 & 0.0047 &0.0282\\ \hline
\end{tabular}
\end{center}}
\end{table*} 

\vspace{.1in}
\noindent {\bf Predicting toxicity matrices using Gene expression data:} We further evaluated the proposed CFM on the toxicity prediction task.  For this experiment, we randomly split the observations of the toxicity matrices into 50\% for training ($341$ elements) and 50\%  for testing ($341$ elements). Then, we used the gene expression matrices $\boldA_1, \boldA_2, \boldA_3$ as side information for predicting the toxicity matrices. More specifically, we designed two types of features from the gene expression data:

\begin{itemize}
\item {\bf Mean of $m$-nearest neighbor similarities:}($x_{\text{mean}}$) We first find the $m$-nearest neighbors of the $i$-th drug target, where the Gaussian kernel is used for similarity computation. Then, we average the similarity of $1, \ldots, m$-th nearest neighbors.  
\item {\bf Standard deviation of $m$-nearest neighbor similarities:}($x_{\text{std}}$) Similarly to the mean feature, we first found $m$-nearest neighbor similarities and then computed that's standard deviation.
\end{itemize}

Then, we used these features as
\begin{align*}
\boldx \!&=\! [\overbrace{0~\cdots 0~\underbrace{1}_{i\text{-th drug}}~0~\cdots 0}^{78}~ \overbrace{0~\underbrace{1}_{k\text{-th sensitivity}}~0}^{3} ~\overbrace{0~\underbrace{1}_{l\text{-th cancer type}}~0}^{3}~\overbrace{x_{\text{mean}}~x_{\text{std}}}^{2} ]^\top \in \mathbbR^{86}, \\
y &= [\boldA_l^{(2)}]_{i,k}.
\end{align*}

We run the prediction experiments on 100 random splits, and report the average RMSE score (Table \ref{tab:result_tox2}). `CFM' is `CFM without any additional features. It is clear that the performance of CFM improves by simply adding manually designed features. Thus, we can improve the prediction performance of CFM by designing new features, and it is useful for various prediction tasks in biology data. 

\begin{table*}[t!]
{
\begin{center}
\caption{Test relative MSE on toxicogenomics data.\label{tab:result_tox2}}
\begin{tabular}{|l||c|c|c|c|c|c|c|c|}
\hline
 & CFM & \multicolumn{3}{c|}{CFM (+mean/std features)} & \multicolumn{3}{c|}{CFM (+mean feature)}\\
 &  & $m=5$ & $m=10$ & $m=15$ & $m=5$ & $m=10$ & $m=15$ \\ \hline
Mean & {0.5624} & {\bf 0.5199} & 0.5207 &  0.5215 & 0.5269 & 0.5234 & 0.5231\\ 
StdError& {0.0501} & 0.0464 &0.0451 & 0.0450 & 0.0466 & 0.0454 &0.0450\\ \hline
\end{tabular}
\end{center}}
\end{table*} 


\section{Conclusion}
We proposed the \emph{convex factorization machine} (CFM), which is a convex variant of factorization machines (FMs).  Specifically, we formulated the CFM optimization problem as a semidefinite program (SDP) and solved it with Hazan's algorithm. A key advantage of the proposed method over FMs is that CFM can find a globally optimal solution, while FMs can get poor locally optimal solutions since they are non-convex approaches. The derived algorithm is simple and can be easily implemented. We also showed the connections between CFM and convex factorization methods and CFM and convex tensor completion methods. Through synthetic and real-world experiments, we showed that the proposed CFM achieves results competitive with state-of-the-art methods. Moreover, for a toxicogenomics prediction task, CFM outperformed a state-of-the-art multi-view tensor factorization method.

In future work, we will extend the proposed method to distributed computation. Another important challenge is to improve the convergence properties of the proposed method. 


\bibliographystyle{unsrt}
\bibliography{main}

\begin{thebibliography}{10}

\bibitem{koren2009matrix}
Yehuda Koren, Robert Bell, and Chris Volinsky.
\newblock Matrix factorization techniques for recommender systems.
\newblock {\em Computer}, (8):30--37, 2009.

\bibitem{wu2007collaborative}
Mingrui Wu.
\newblock Collaborative filtering via ensembles of matrix factorizations.
\newblock In {\em KDD}, 2007.

\bibitem{xu2003document}
Wei Xu, Xin Liu, and Yihong Gong.
\newblock Document clustering based on non-negative matrix factorization.
\newblock In {\em SIGIR}, 2003.

\bibitem{pan2010transfer}
Weike Pan, Evan~Wei Xiang, Nathan~Nan Liu, and Qiang Yang.
\newblock Transfer learning in collaborative filtering for sparsity reduction.
\newblock In {\em AAAI}, 2010.

\bibitem{xu2010protein}
Qian Xu, Evan~Wei Xiang, and Qiang Yang.
\newblock Protein-protein interaction prediction via collective matrix
  factorization.
\newblock In {\em BIBM}, 2010.

\bibitem{hong2013co}
Liangjie Hong, Aziz~S Doumith, and Brian~D Davison.
\newblock Co-factorization machines: modeling user interests and predicting
  individual decisions in twitter.
\newblock In {\em WSDM}, 2013.

\bibitem{virtanen2011bayesian}
Seppo Virtanen, Arto Klami, and Samuel Kaski.
\newblock Bayesian {CCA} via group sparsity.
\newblock In {\em ICML}, 2011.

\bibitem{lian2015integrating}
Wenzhao Lian, Piyush Rai, Esther Salazar, and Lawrence Carin.
\newblock Integrating features and similarities: Flexible models for
  heterogeneous multiview data.
\newblock In {\em AAAI}, 2015.

\bibitem{gunasekarconsistent}
Suriya Gunasekar, Makoto Yamada, Dawei Yin, and Yi~Chang.
\newblock Consistent collective matrix completion under joint low rank
  structure.
\newblock In {\em AISTATS}, 2015.

\bibitem{yan2015scalable}
Yan Yan, Mingkui Tan, Ivor Tsang, Yi~Yang, Chengqi Zhang, and Qinfeng Shi.
\newblock Scalable maximum margin matrix factorization by active {R}iemannian
  subspace search.
\newblock In {\em IJCAI}, 2015.

\bibitem{rendle2010factorization}
Steffen Rendle.
\newblock Factorization machines.
\newblock In {\em ICDM}, 2010.

\bibitem{rendle2012factorization}
Steffen Rendle.
\newblock Factorization machines with lib{F}m.
\newblock {\em ACM Transactions on Intelligent Systems and Technology (TIST)},
  3(3):57, 2012.

\bibitem{rendle2013scaling}
Steffen Rendle.
\newblock Scaling factorization machines to relational data.
\newblock In {\em VLDB}, volume~6, pages 337--348, 2013.

\bibitem{hazan2008sparse}
Elad Hazan.
\newblock Sparse approximate solutions to semidefinite programs.
\newblock In {\em LATIN 2008: Theoretical Informatics}. 2008.

\bibitem{frank1956algorithm}
Marguerite Frank and Philip Wolfe.
\newblock An algorithm for quadratic programming.
\newblock {\em Naval {R}esearch {L}ogistics {Q}uarterly}, 3(1-2):95--110, 1956.

\bibitem{jaggi2013revisiting}
Martin Jaggi.
\newblock Revisiting frank-wolfe: Projection-free sparse convex optimization.
\newblock In {\em ICML}, 2013.

\bibitem{khan2014bayesian}
Suleiman~A Khan and Samuel Kaski.
\newblock Bayesian multi-view tensor factorization.
\newblock In {\em ECML}. 2014.

\bibitem{jaggi2010simple}
Martin Jaggi and Marek Sulovsky.
\newblock A simple algorithm for nuclear norm regularized problems.
\newblock In {\em ICML}, 2010.

\bibitem{tucker1966some}
Ledyard~R Tucker.
\newblock Some mathematical notes on three-mode factor analysis.
\newblock {\em Psychometrika}, 31(3):279--311, 1966.

\bibitem{tomioka2010estimation}
Ryota Tomioka, Kohei Hayashi, and Hisashi Kashima.
\newblock Estimation of low-rank tensors via convex optimization.
\newblock {\em arXiv preprint arXiv:1010.0789}, 2010.

\bibitem{tomioka2013convex}
Ryota Tomioka and Taiji Suzuki.
\newblock Convex tensor decomposition via structured schatten norm
  regularization.
\newblock In {\em NIPS}, 2013.

\bibitem{combettes2005signal}
Patrick~L Combettes and Val{\'e}rie~R Wajs.
\newblock Signal recovery by proximal forward-backward splitting.
\newblock {\em Multiscale Modeling \& Simulation}, 4(4):1168--1200, 2005.

\bibitem{choi2014dfacto}
Joon~Hee Choi and S~Vishwanathan.
\newblock Dfacto: Distributed factorization of tensors.
\newblock In {\em NIPS}, 2014.

\bibitem{convex_fm}
Mathieu Blondel, Akinori Fujino, and Naonori Ueda.
\newblock Convex factorization machines.
\newblock In {\em ECMLPKDD}, 2015.

\bibitem{fazel2001rank}
Maryam Fazel, Haitham Hindi, and Stephen~P Boyd.
\newblock A rank minimization heuristic with application to minimum order
  system approximation.
\newblock In {\em ACC}, 2001.

\bibitem{candes2009exact}
Emmanuel~J Cand{\`e}s and Benjamin Recht.
\newblock Exact matrix completion via convex optimization.
\newblock {\em Foundations of Computational mathematics}, 9(6):717--772, 2009.

\bibitem{ji2009accelerated}
Shuiwang Ji and Jieping Ye.
\newblock An accelerated gradient method for trace norm minimization.
\newblock In {\em ICML}, 2009.

\bibitem{bach2008convex}
Francis Bach, Julien Mairal, and Jean Ponce.
\newblock Convex sparse matrix factorizations.
\newblock {\em arXiv preprint arXiv:0812.1869}, 2008.

\bibitem{toh2010accelerated}
Kim-Chuan Toh and Sangwoon Yun.
\newblock An accelerated proximal gradient algorithm for nuclear norm
  regularized linear least squares problems.
\newblock {\em Pacific Journal of Optimization}, 6(615-640):15, 2010.

\bibitem{tomioka2011statistical}
Ryota Tomioka, Taiji Suzuki, Kohei Hayashi, and Hisashi Kashima.
\newblock Statistical performance of convex tensor decomposition.
\newblock In {\em NIPS}, 2011.

\bibitem{liu2012implementable}
Yong-Jin Liu, Defeng Sun, and Kim-Chuan Toh.
\newblock An implementable proximal point algorithmic framework for nuclear
  norm minimization.
\newblock {\em Mathematical {P}programming}, 133(1-2):399--436, 2012.

\bibitem{mazumder2010spectral}
Rahul Mazumder, Trevor Hastie, and Robert Tibshirani.
\newblock Spectral regularization algorithms for learning large incomplete
  matrices.
\newblock {\em JMLR}, 11:2287--2322, 2010.

\bibitem{cai2010singular}
Jian-Feng Cai, Emmanuel~J Cand{\`e}s, and Zuowei Shen.
\newblock A singular value thresholding algorithm for matrix completion.
\newblock {\em SIAM Journal on Optimization}, 20(4):1956--1982, 2010.

\bibitem{shalev2011large}
Shai Shalev-Shwartz, Alon Gonen, and Ohad Shamir.
\newblock Large-scale convex minimization with a low-rank constraint.
\newblock In {\em ICML}, 2011.

\bibitem{hsieh2014nuclear}
Cho-Jui Hsieh and Peder Olsen.
\newblock Nuclear norm minimization via active subspace selection.
\newblock In {\em ICML}, 2014.

\bibitem{singh2008relational}
Ajit~P Singh and Geoffrey~J Gordon.
\newblock Relational learning via collective matrix factorization.
\newblock In {\em KDD}, 2008.

\bibitem{bouchard2013convex}
Guillaume Bouchard, Dawei Yin, and Shengbo Guo.
\newblock Convex collective matrix factorization.
\newblock In {\em AISTATS}, 2013.

\bibitem{miller2003movielens}
Bradley~N Miller, Istvan Albert, Shyong~K Lam, Joseph~A Konstan, and John
  Riedl.
\newblock Movielens unplugged: {E}xperiences with an occasionally connected
  recommender system.
\newblock In {\em IUI}, 2003.

\bibitem{lamb2006cmap}
Justin Lamb et~al.
\newblock The connectivity map: Using gene-expression signatures to connect
  small molecules, genes, and disease.
\newblock {\em Science}, 313(5795):1929--1935, 2006.

\bibitem{shoemaker2006nci60}
Robert~H Shoemaker.
\newblock The {NCI}60 human tumour cell line anticancer drug screen.
\newblock {\em Nature Reviews Cancer}, 6(10):813--823, 2006.

\bibitem{morup2009automatic}
Morten M{\o}rup and Lars~Kai Hansen.
\newblock Automatic relevance determination for multi-way models.
\newblock {\em Journal of Chemometrics}, 23(7-8):352--363, 2009.

\bibitem{carroll1970analysis}
J~Douglas Carroll and Jih-Jie Chang.
\newblock Analysis of individual differences in multidimensional scaling via an
  n-way generalization of “eckart-young” decomposition.
\newblock {\em Psychometrika}, 35(3):283--319, 1970.

\bibitem{virtanen2012bayesian}
Seppo Virtanen, Arto Klami, Suleiman~A Khan, and Samuel Kaski.
\newblock Bayesian group factor analysis.
\newblock In {\em AISTATS}, 2012.

\end{thebibliography}

\end{document}